\documentclass[sigconf]{acmart}

\makeatletter
\def\@ACM@checkaffil{
    \if@ACM@instpresent\else
    \ClassWarningNoLine{\@classname}{No institution present for an affiliation}%
    \fi
    \if@ACM@citypresent\else
    \ClassWarningNoLine{\@classname}{No city present for an affiliation}%
    \fi
    \if@ACM@countrypresent\else
        \ClassWarningNoLine{\@classname}{No country present for an affiliation}%
    \fi
}
\makeatother

\usepackage{epsfig}
\usepackage{graphicx}
\usepackage{amsmath}
\usepackage{helvet}  
\usepackage{courier}  
\usepackage{url}  
\usepackage{xcolor}
\usepackage{multirow}
\usepackage{subcaption}
\usepackage{array}

\usepackage{color, colortbl}
\definecolor{mygray}{gray}{0.9}

\newcolumntype{P}[1]{>{\centering\arraybackslash}p{#1}}
\acmConference[SIGIR'24]{ACM SIGIR Conference}{Jul.}{U.S.}
\newcommand{\blfootnote}[1]{%
  \begingroup
  \renewcommand\thefootnote{}\footnote{#1}%
  \addtocounter{footnote}{-1}%
  \endgroup
}

\def\BibTeX{{\rm B\kern-.05em{\sc i\kern-.025em b}\kern-.08emT\kern-.1667em\lower.7ex\hbox{E}\kern-.125emX}}
    
\begin{document}
\acmSubmissionID{4}
%

\title{A Novel Evaluation Framework for Image2Text Generation}


\author{Jia-Hong Huang$^*$}
\affiliation{\institution{\small{University of Amsterdam, Netherlands} \\ \tt{j.huang@uva.nl}}}

\author{Hongyi Zhu$^*$}
\affiliation{\institution{\small{University of Amsterdam, Netherlands} \\ \tt{h.zhu@uva.nl}}}

\author{Yixian Shen}
\affiliation{\institution{\small{University of Amsterdam, Netherlands} \\ \tt{y.shen@uva.nl}}}

\author{Stevan Rudinac}
\affiliation{\institution{\small{University of Amsterdam, Netherlands} \\ \tt{s.rudinac@uva.nl}}}

\author{Alessio M. Pacces}
\affiliation{\institution{\small{University of Amsterdam, Netherlands} \\ \tt{a.m.pacces@uva.nl}}}

\author{Evangelos Kanoulas}
\affiliation{\institution{\small{University of Amsterdam, Netherlands} \\ \tt{E.Kanoulas@uva.nl}}}

%
\begin{abstract}
Evaluating the quality of automatically generated image descriptions is challenging, requiring metrics that capture various aspects such as grammaticality, coverage, correctness, and truthfulness. While human evaluation offers valuable insights, its cost and time-consuming nature pose limitations. Existing automated metrics like BLEU, ROUGE, METEOR, and CIDEr aim to bridge this gap but often show weak correlations with human judgment. We address this challenge by introducing a novel evaluation framework rooted in a modern large language model (LLM), such as GPT-4 or Gemini, capable of image generation.
In our proposed framework, we begin by feeding an input image into a designated image captioning model, chosen for evaluation, to generate a textual description. Using this description, an LLM then creates a new image. By extracting features from both the original and LLM-created images, we measure their similarity using a designated similarity metric. A high similarity score suggests that the image captioning model has accurately generated textual descriptions, while a low similarity score indicates discrepancies, revealing potential shortcomings in the model's performance. Human-annotated reference captions are not required in our proposed evaluation framework, which serves as a valuable tool for evaluating the effectiveness of image captioning models. Its efficacy is confirmed through human evaluation. 
\end{abstract}


%
%


%

%


%
\maketitle
\blfootnote{$*$ Equal contribution.}

\vspace{-0.8cm}
\section{Introduction}

The evaluation of sentences generated through automated methods remains a formidable challenge in the realm of image captioning. Current metrics for evaluating image descriptions aim to gauge multiple desirable attributes, such as grammaticality, covering crucial aspects, correctness, truthfulness, and more. Human evaluation plays a pivotal role in quantifying these properties, utilizing separate Likert scales or pairwise scales \cite{mitchell2012midge,rohrbach2013translating,yang2011corpus,elliott2013image,yatskar2014see,linkert1932technique}. 
However, due to the expensive, challenging-to-reproduce, and time-consuming nature of human studies, there is a growing need for automated evaluation measures. For practical utility, these automated metrics should align closely with human judgment. Therefore, the challenge in designing such an automatic metric lies in integrating the aforementioned diverse evaluations attributes into a unified measure of sentence quality.

\begin{figure*}[t!]
\begin{center}
\includegraphics[width=1.0\linewidth]{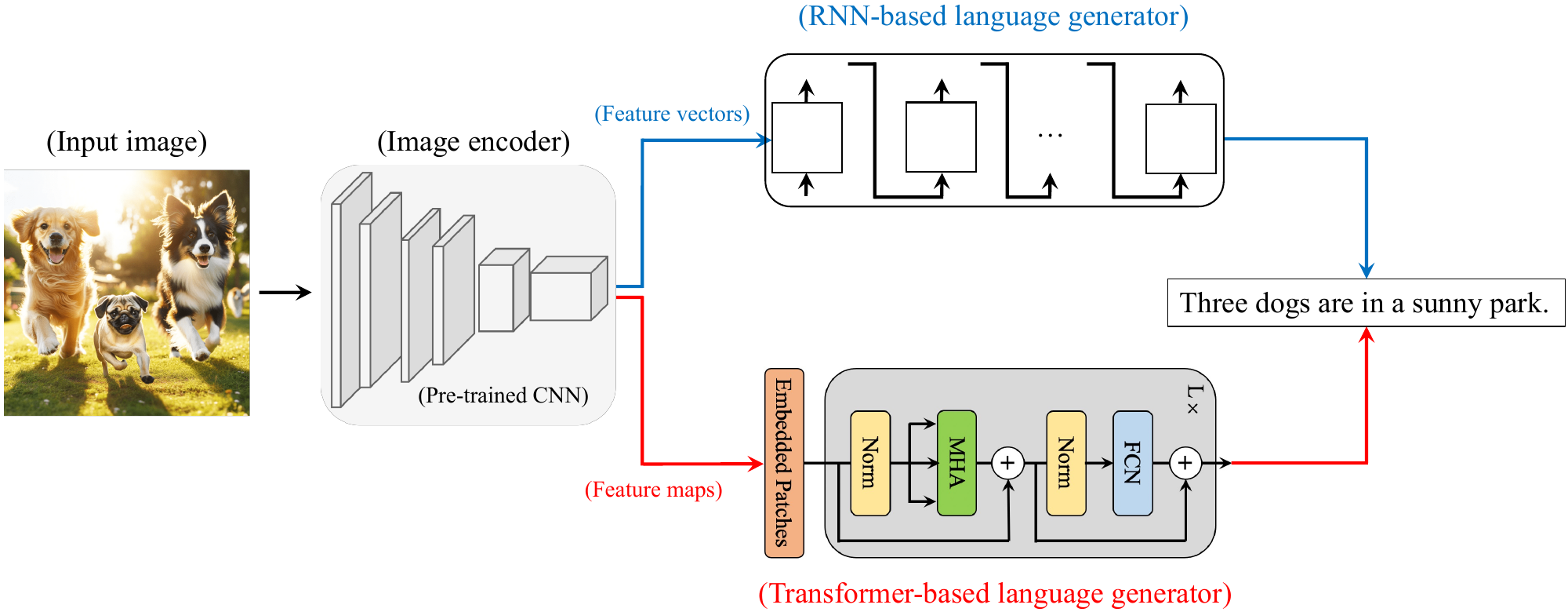}
\end{center}
\vspace{-0.4cm}
   \caption{Flowchart for image captioning. Existing image captioning architectures can be broadly categorized into two groups: those based on the recurrent neural network (RNN) and those based on the transformer architecture. To aid comprehension, we represent RNN-based methods with blue paths and transformer-based approaches with red paths. The process involves feeding an input image through an image encoder for feature extraction, followed by a language generator to produce text-based descriptions using the extracted image features.}
\vspace{-0.4cm}
\label{fig:figure1}
\end{figure*}

Several automated metrics, including BLEU \cite{papineni2002bleu}, ROUGE \cite{lin2004rouge}, METEOR \cite{elliott2013image}, CIDEr \cite{vedantam2015cider}, and more, have been introduced to assess image descriptions generated by automated approaches. BLEU, initially designed for machine translation, relies on precision, while ROUGE, originating from the summarization community, is a recall-based metric. METEOR is tailored for assessing the overall quality of image descriptions. Nonetheless, research has indicated a weak correlation between these metrics and human judgment \cite{bybaby,elliott2013image,callison2006re,hodosh2013framing}.
In contrast, the consensus-based metric CIDEr measures the similarity between a generated sentence and a set of ground truth sentences authored by humans, demonstrating high agreement with human consensus. However, preparing a set of ground truth sentences in advance is a prerequisite for CIDEr. If the quantity of human-authored ground truth sentences is insufficient, CIDEr may struggle to effectively evaluate image descriptions \cite{vedantam2015cider}. A similar limitation is observed in the CLAIR method \cite{chan2023clair} and other aforementioned approaches. Some metrics involve caption ranking \cite{hodosh2013framing} but are limited in evaluating novel image descriptions.

In addressing the above challenge, we present a novel framework for evaluating image descriptions. This framework is rooted in the utilization of a modern LLM approach, e.g., GPT-4 \cite{brown2020language} or Gemini \cite{team2023gemini}, capable of generating images.
The advancement of LLMs \cite{wu2023brief,vaswani2017attention}, exemplified by models like GPT-4, empowers us to provide textual descriptions, i.e., prompt, for generating images that closely correspond and align with the semantic meaning conveyed in the given text. 
The underlying design philosophy of the proposed framework hinges on the idea that if an image captioning model is validated as effective, the generated image description by the model should be sufficiently accurate to reconstruct the same or a highly similar image compared to the original input image, relying on LLMs. The ongoing evolution of LLM technology forms the bedrock of the proposed framework.

Starting with the definition of the image captioning task, as illustrated in Figure \ref{fig:figure1}, our proposed framework begins by taking an image as input. Subsequently, this input undergoes processing through a given image captioning model, generating a textual description for the initial image. Following this, a given LLM, such as GPT-4, is employed to generate an image based on the textual description. Then, we extract the image features from both the original input image and the LLM-generated image, and assess their similarity using the cosine similarity metric. 
It is worth noting that human-annotated reference captions are not needed in our proposed evaluation framework.
In the proposed evaluation framework, a high cosine similarity score is anticipated if the generated text-based description is of sufficient quality, signifying that the LLM can accurately reproduce an image highly similar to the original input. Conversely, if the generated text-based description lacks accuracy, the image produced by the LLM will deviate from the original input image and lead to a low cosine similarity score. 
This incongruity suggests the suboptimal performance of the image captioning model. Consequently, the proposed framework proves valuable for evaluating the efficacy of a given image captioning model.

\noindent\vspace{+3pt}
The main contributions of this work are summarized as follows:
\vspace{-1cm}
\begin{itemize}
    \item \textbf{Innovative Framework for Image Captioning Model Evaluation:} We present a novel framework that relies on the utilization of an LLM, such as GPT-4 or Gemini, to evaluate the quality of image descriptions generated by an image captioning model. The proposed evaluation framework does not necessitate human-annotated reference captions.

    \item \textbf{Human Evaluation of the Framework:} To verify the effectiveness of our evaluation framework, we introduce a human-annotated dataset and conduct human evaluations.

    \item  \textbf{Comprehensive Experiments on Established Datasets:} We perform extensive experiments to demonstrate the efficacy of the proposed evaluation framework using widely-used image captioning datasets.
\end{itemize}

\section{Related Work}
In this section, we begin by reviewing existing related literature, covering topics such as the existing image captioning methods, the evolution of automated metrics, and the latest advancements in LLM technology.

\subsection{Image Captioning Methods}
The encoder-decoder network architecture has become a cornerstone in the field of image captioning, as evidenced by various studies \cite{xiao2019deep,vinyals2015show,jun2020t,huang2017robustness,huang2017vqabq,huang2019assessing,huang2023improving,huang2017robustnessMS,huang2019novel_1,huang2020query,huang2021gpt2mvs,huang2022causal,wang2024ada,huang2023causalainer,huang2023query,huang2023conditional,hu2019silco}. Typically, these networks employ a CNN as the encoder for extracting global image features, and an RNN as the decoder for generating word sequences. \cite{mao2016generation} introduces a method for generating referring expressions, which are descriptions for specific objects or regions within an image. In \cite{wang2016image}, the bidirectional LSTM-based method for image captioning takes advantage of both past and future information to learn long-term visual-language interactions.
Attention mechanisms have significantly enhanced the performance of image captioning models. \cite{pedersoli2017areas} introduces an area-based attention model that predicts the next word and the corresponding image regions at each RNN timestep. While these advancements represent significant strides, they predominantly focus on single-image based description generation. However, certain abstract concepts or descriptions might not be fully captured using only image data \cite{laserson2018textray,jing2017automatic}. \cite{huang2022non,huang2021deepopht,huang2021contextualized,huang2024multi,yang2018novel,di2021dawn,huang2021deep,liu2018synthesizing,huang2021longer,wu2023expert,huck2018auto} have explored the use of expert-defined keyword sequences to augment model capabilities in generating more accurate and contextually relevant descriptions.
Recent advancements have also explored transformer-based architectures, such as Vision Transformers (ViT), which have shown promise in capturing finer details and global context in images for caption generation \cite{dosovitskiy2020image}. Furthermore, the integration of multimodal learning approaches, where models are trained on both visual and textual data, has led to large improvements in generating contextually richer and more nuanced image descriptions \cite{lu2019vilbert}.

The domain of medical image captioning has witnessed significant advancements, particularly through methods that meld human expertise with algorithmic prowess. \cite{li2018hybrid} has developed a Hybrid Retrieval-Generation Reinforced Agent, which integrates human prior knowledge with AI-based caption generation for medical images. This agent alternates between a generative module and a retrieval mechanism that utilizes a template database reflecting human expertise, thereby producing multi-faceted, sequential sentences. \cite{jing2017automatic} has contributed to this field with a multi-task learning framework that simultaneously predicts tags and generates captions. Their method, which focuses on abnormal areas in chest radiology images using an attention mechanism and a hierarchical LSTM, offers detailed descriptions. 
These methods primarily focus on generating reports for chest radiology images, which are structurally different in terms of object size and detail compared to retinal images \cite{laserson2018textray,tierney2020comparative,huang2021deepopht}. Additionally, the color features in chest radiology and retinal images differ significantly, with the former being predominantly grey-scale and the latter being colorful \cite{laserson2018textray,huang2021deepopht}. Most existing methods rely primarily on the image input for caption generation.
Recent advancements also include the enhancement of the CNN-RNN framework with the TransFuser model \cite{huang2022non}. This model adeptly combines features from different modalities and addresses the challenge of incorporating unordered keyword sequences with visual inputs, minimizing information loss \cite{huang2022non}. This development represents a significant stride in medical image captioning, reflecting the growing complexity and capability of these methods.
Further progress in deep learning, particularly the application of ViTs, has offered promising results in medical imaging \cite{chen2021vit}. ViTs excel in capturing intricate details and providing a broader context for more accurate medical image analysis and caption generation.

The evaluation framework proposed in this paper is versatile and capable of assessing any existing image captioning approaches. 

\vspace{-0.4cm}
\subsection{Automatic Metrics for Image Captioning}
The evolution of image captioning has been significantly influenced by the development and application of automatic metrics for evaluating caption quality \cite{papineni2002bleu,lin2004rouge,banerjee2005meteor,vedantam2015cider,anderson2016spice, sharma2017relevance}. These metrics guide the training of captioning models and provide a scalable means for performance assessment. The BLEU score, a pioneering metric by \cite{papineni2002bleu}, gauges n-gram precision in generated text against a reference. ROUGE, developed by \cite{lin2004rouge}, emphasizes recall through the overlap of N-grams and longest common subsequences.
Subsequent innovations introduced refined approaches. METEOR, by \cite{banerjee2005meteor}, aligns more closely with human judgment by incorporating synonym matching and stemming.In \cite{vedantam2015cider}, the CIDEr metric, specifically designed for image captioning, assesses the similarity of generated captions to a set of reference captions. The SPICE metric by \cite{anderson2016spice} evaluates semantic content and the depiction of objects, attributes, and relationships. Additionally, the NLG-Eval toolkit by \cite{sharma2017relevance} provides a comprehensive suite of metrics for a more holistic evaluation of natural language generation.
However, these metrics have limitations. Metrics like BLEU and ROUGE often fail to capture the contextual nuances of captions \cite{papineni2002bleu,lin2004rouge}. The challenge of evaluating creativity and novelty in caption generation is also evident, as automated metrics may penalize deviations from standard references \cite{vedantam2015cider,anderson2016spice}. Recently, advancements like BERTScore \cite{zhang2019bertscore} and CLIPScore \cite{hessel2021clipscore}, which utilize contextual embeddings and visual-textual alignment, respectively, have been proposed to address these challenges.

In this study, human evaluation is employed to validate the effectiveness of the proposed evaluation framework.

\begin{figure*}[ht!]
  \includegraphics[width=\textwidth]{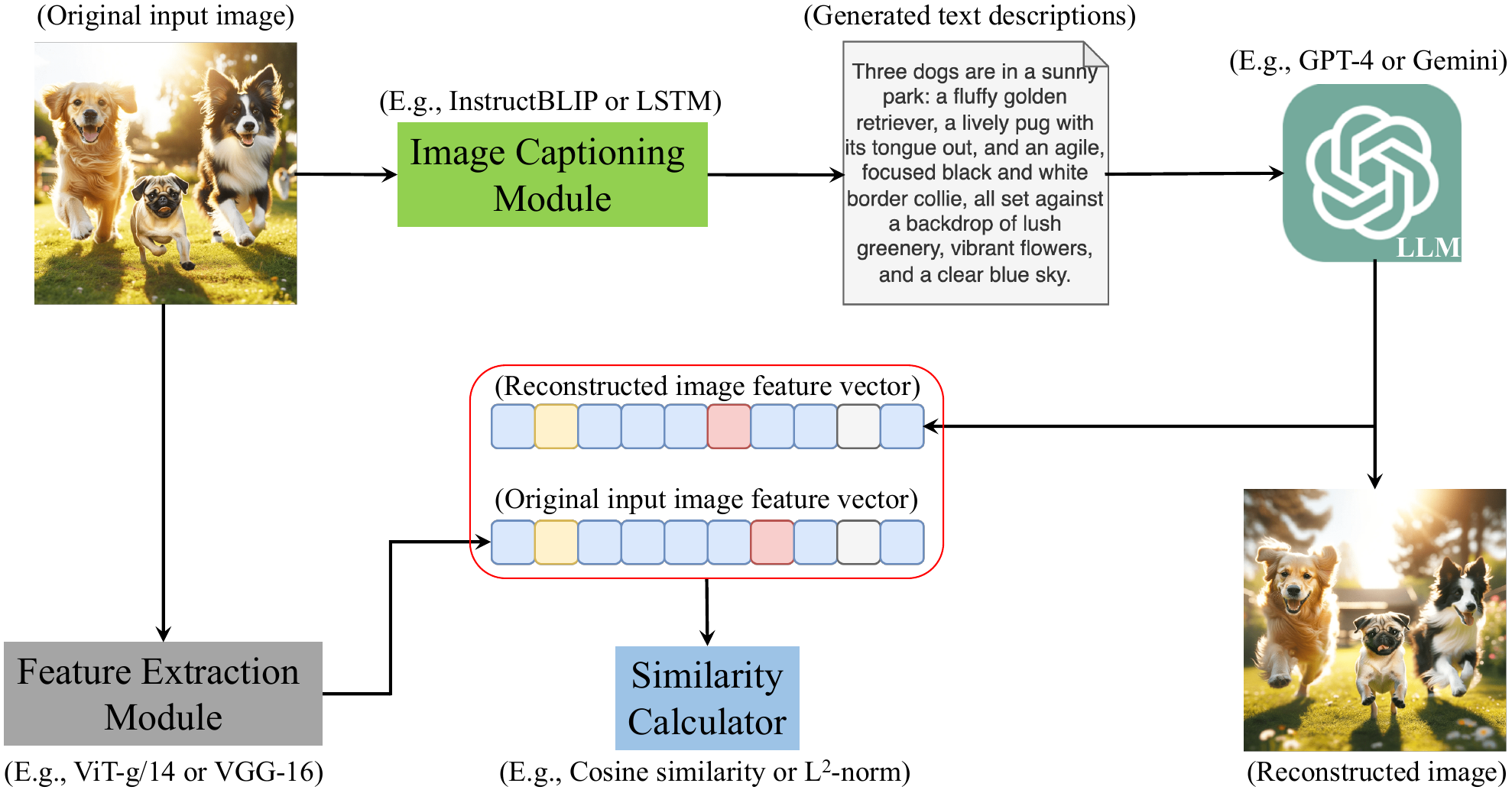}
  \vspace{-0.7cm}
  \caption{
  Flowchart of the proposed evaluation framework. The proposed framework consists of four main components: an image captioning module, an image feature extractor, a large language model (LLM), and a similarity calculator. The image captioning module employs a chosen model to process an input image and generate textual descriptions. The image feature extractor is tasked with extracting features from the input image. The LLM utilizes the text descriptions produced by the image captioning model to generate the corresponding image. Finally, the similarity calculator computes the similarity between the features of the input image and the image generated by the LLM.
  }
  \label{fig:figure2}
  \vspace{-0.4cm}
\end{figure*}

\subsection{Large Language Models}
The advent of LLMs has significantly reshaped the landscape of natural language processing (NLP) and Artificial Intelligence (AI) \cite{zhu2024enhancing,huang2024optimizing,zhang2024comparative,zhang2024beyond,zhang2024towards,zhang2024qfmts}. Pioneering models such as GPT, developed by \cite{brown2020language}, and BERT by \cite{devlin2018bert}, have marked critical milestones in this evolution. These models, characterized by their vast number of parameters and advanced deep learning architectures, have enhanced the capacity to understand and generate human language, excelling in diverse tasks like translation, summarization, and question-answering \cite{devlin2018bert, radford2019language}.
The efficacy of LLMs such as GPT, which utilizes a transformer-based architecture, stems from their comprehensive training across a broad spectrum of internet text, enabling the generation of coherent and contextually pertinent language \cite{radford2019language}. BERT's introduction of bidirectional transformers has revolutionized pre-training in language understanding, showing remarkable efficiency in tasks requiring intricate contextual comprehension \cite{devlin2018bert}. The incorporation of attention mechanisms, as conceptualized by \cite{vaswani2017attention}, has further refined these models' ability for nuanced understanding and text generation.
In the realm of image captioning, the deployment of LLMs like GPT-3 has brought transformative changes. GPT-3's adeptness in image captioning tasks is a testament to its sophisticated transformer-based architecture and comprehensive training on a wide array of internet text. This extensive training enables GPT-3 to intricately understand and generate content that accurately aligns with both textual and visual contexts, producing coherent, contextually relevant, and detailed image descriptions \cite{radford2019language}. The fusion of LLMs with advanced computer vision techniques has been a significant leap forward, leading to the development of more sophisticated systems. These systems are now better equipped to interpret and describe complex visual data with greater accuracy and nuance \cite{jia2021scaling}. This integration highlights the evolving capability of AI to understand and convey the subtleties of visual information, mirroring a more human-like perception and articulation of images. This advancement in image captioning technology is pivotal in enhancing how machines process and narrate visual data, bridging the gap between visual perception and linguistic expression.
Furthermore, the use of LLMs goes beyond generating captions to evaluating their quality. A notable method in this regard is CLAIR \cite{chan2023clair}, which leverages zero-shot language modeling to assess caption quality. CLAIR shows a stronger correlation with human judgment compared to traditional metrics like BLEU, ROUGE, METEOR, and CIDEr. By soliciting an LLM to rate how likely a candidate caption accurately describes an image relative to a set of reference captions, CLAIR outperforms language-only measures, approaching human-level correlation. However, CLAIR requires a set of human-annotated reference captions to function, without which it cannot be applied.

In this work, the proposed approach leverages modern LLMs like GPT-4 for an innovative and comprehensive evaluation. We use LLMs to reverse-engineer the image captioning process, generating images from textual descriptions to assess caption accuracy. This method offers a unique advantage in evaluating the semantic richness and contextual relevance of captions. By comparing the generated images with the original, our approach provides a direct, visual assessment of caption quality, moving beyond mere textual analysis. This method not only aligns with human perception but also embraces the creativity and diversity inherent in image captioning, offering a more rounded and practical evaluation framework.



\vspace{-0.7cm}
\section{Methodology}
The proposed evaluation framework comprises several key components: an image captioning module, an LLM-based text-to-image generator, an image feature extraction module, and a similarity calculator, as depicted in Figure \ref{fig:figure2}. Each of these components will be introduced in detail in the following subsections. Furthermore, to ensure the validity of the evaluation results based on our framework—specifically, their alignment with human judgment—we introduce a human-annotated image captioning dataset to validate the effectiveness of the proposed framework.

\subsection{Image Captioning Module}
The module incorporates an image captioning model, which will undergo evaluation using the proposed framework. This module takes an image as input and generates a text-based description as output. To facilitate user comprehension of the proposed evaluation framework, we utilize the InstructBLIP model \cite{dai2305instructblip} as an illustrative example in Section~\ref{exp:Experiments and Analysis}. This demonstration showcases the entire process of leveraging the proposed framework to evaluate a given image captioning model, making it easily understandable for users.

\subsection{LLM-based Text-to-Image Generator}
Numerous studies \cite{brown2020language,team2023gemini,wu2023brief} have demonstrated the proficiency of LLM-based image generators, exemplified by models like GPT-4, in producing high-quality images that closely align with the semantic meaning of provided text-based prompts. Specifically, DALL-E, functioning as an image generation model within GPT-4, a variant of GPT-3 boasting 12 billion parameters, is engineered by OpenAI to generate images based on textual descriptions, drawing from a dataset comprising text-image pairs. Its versatile capabilities include crafting anthropomorphized versions of animals and objects, seamlessly combining unrelated concepts, rendering text, and applying transformations to existing images. 
In the context of the proposed framework, the LLM-based image generator utilizes the text-based image description generated by a preceding image captioning model. If the image captioning model performs well, generating a high-quality and accurate image description, the LLM-based image generator subsequently creates an image that is similar to the original input image. This connection highlights the interplay between effective image captioning and the generation of corresponding images by the LLM-based approach. 

\subsection{Image Feature Extraction Module}

The image feature extraction module primarily consists of a pre-trained image encoder. This module takes an image as input and produces a feature vector representing the input image as output. To enhance user understanding of the proposed evaluation framework, we employ ViT-g/14 \cite{fang2023eva} as a demonstrative example for image feature extraction in Section \ref{exp:Experiments and Analysis}.
ViT-g/14 is a vanilla ViT pre-trained for reconstructing masked-out image-text aligned vision features conditioned on visible image patches. Through this pretext task, the model efficiently scales up to one billion parameters, achieving notable performance across various vision downstream tasks, including image recognition, video action recognition, object detection, instance segmentation, and semantic segmentation, all without extensive supervised training.
This demonstration in Section \ref{exp:Experiments and Analysis} highlights the complete process, encompassing image feature extraction for calculating similarity scores between the input and generated images. It illustrates how the proposed framework can be leveraged to assess a given image captioning model, providing users with a clear understanding. It is worth noting that the image feature extractor can be substituted with other pre-trained CNNs, such as VGG-16 \cite{simonyan2014very} or ResNet-52 \cite{he2016deep}.

\subsection{Similarity Calculator}
\label{metric:cosine}
Cosine similarity, as defined in Equation~(\ref{eq:cossim}), serves as a metric for quantifying the similarity between two vectors in a multi-dimensional space. It evaluates the cosine of the angle between these vectors, offering insight into their degree of similarity or dissimilarity.
The advantage of cosine similarity lies in its ability to assess directional similarity rather than magnitude, rendering it robust against variations in scale and orientation. This characteristic makes it a widely adopted metric in diverse domains, including image processing and NLP. In these fields, cosine similarity is frequently employed to assess the similarity between images, documents, or sentences represented as vectors in high-dimensional spaces. The cosine similarity value $\text{CosSim}(\cdot~, \cdot) \in [-1, 1]$, where a value of $1$ signifies that the vectors are identical, $0$ indicates orthogonality (i.e., no similarity), and $-1$ indicates complete dissimilarity or opposition.
\begin{equation}
    \text{CosSim}(\mathbf{i_{o}}, \mathbf{i_{g}}) = \frac{\mathbf{i_{o}} \cdot \mathbf{i_{g}}}{\|\mathbf{i_{o}}\| \|\mathbf{i_{g}}\|},
\label{eq:cossim}
\end{equation}
where $\mathbf{i_{o}} \cdot \mathbf{i_{g}}$ denotes the dot product (also known as the inner product) of the original input image feature vector $\mathbf{i_{o}}$ and the LLM-generated image feature vector $\mathbf{i_{g}}$. $\|\mathbf{i_{o}}\|$ and $\|\mathbf{i_{g}}\|$ represent the Euclidean norms (also known as the magnitudes or lengths) of vectors $\mathbf{i_{o}}$ and $\mathbf{i_{g}}$, respectively. In words, cosine similarity measures the cosine of the angle between two vectors, which represents their similarity in direction and magnitude.

\subsection{Human-annotated Image Captioning Dataset}
\label{dataset:dataset}
\begin{figure*}[t!]
\begin{center}
\includegraphics[width=1.0\linewidth]{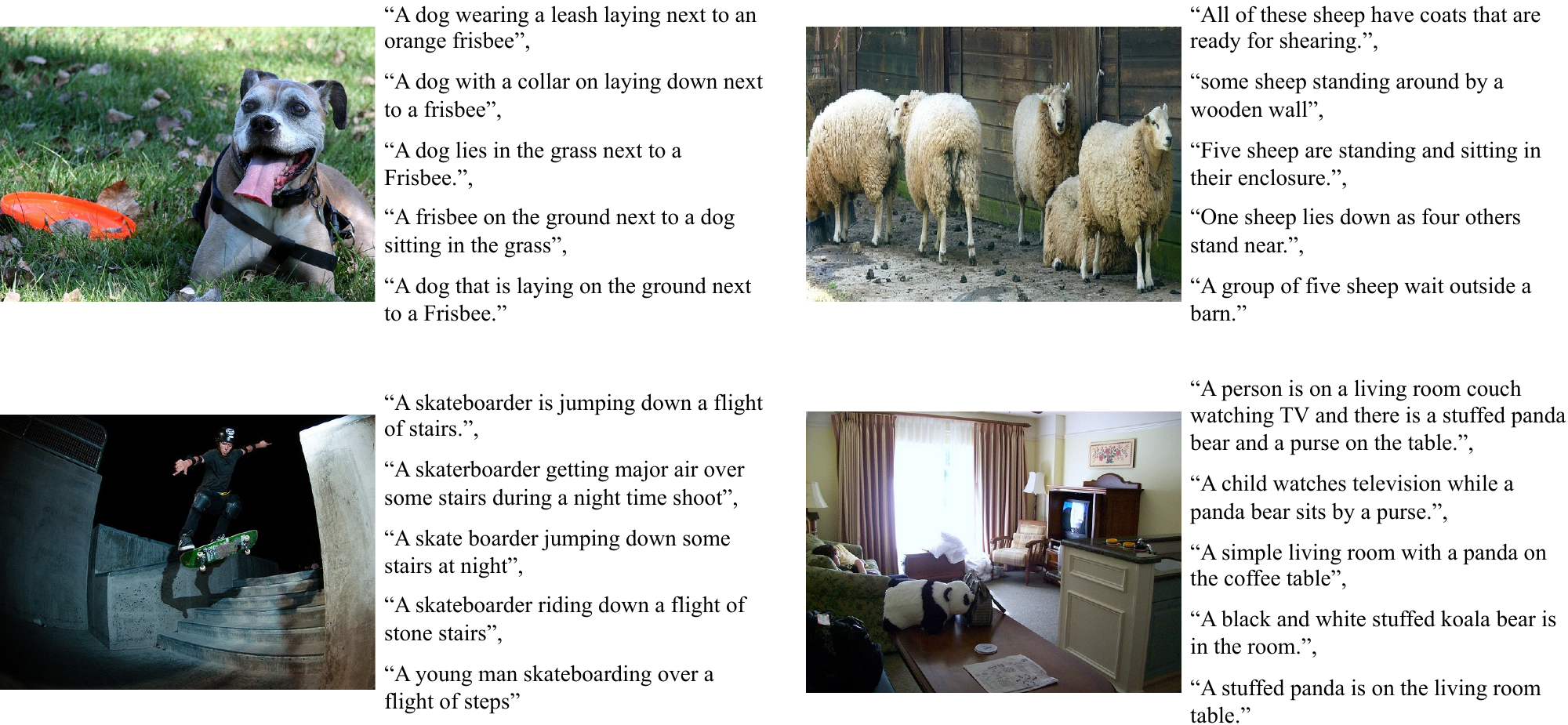}
\end{center}
\vspace{-0.4cm}
   \caption{Dataset examples. To provide a clearer insight into the introduced human-annotated dataset, we have randomly selected four examples for illustrative purposes. Each image in the dataset is accompanied by five human-annotated descriptions that vividly depict the content of the image.}
\vspace{-0.4cm}
\label{fig:figure3}
\end{figure*}

The Microsoft Common Objects in Context (MSCOCO) dataset is a comprehensive resource widely used across various image recognition tasks including object detection, segmentation, and captioning. Originally, the MSCOCO Captions dataset comprised over $330,000$ images, each meticulously annotated with $80$ object categories. Notably, both the training and validation sets feature each image accompanied by five distinct human-generated captions. This dataset holds significant importance within the realm of computer vision research, serving as a cornerstone for the development and evaluation of numerous state-of-the-art object detection and segmentation models. In our study, we enhance the existing MSCOCO Caption dataset by incorporating an additional $30,000$ human-annotated image-description pairs. This augmented dataset serves as the basis for evaluating the alignment of our proposed evaluation method with human-annotated image descriptions. To aid in understanding the dataset, several examples from the dataset are provided in Figure \ref{fig:figure3}.

\section{Experiments and Analysis}
\label{exp:Experiments and Analysis}
In this section, our goal is to evaluate the effectiveness of the proposed evaluation framework designed for image captioning models. To achieve this, we will validate our framework using both the widely adopted human-annotated image captioning datasets and our newly introduced dataset, the details of which are outlined in the Section \ref{dataset:dataset}. Since all datasets have undergone human annotation, our primary objective in this assessment is to ascertain whether the evaluation results obtained through our proposed framework align with human consensus or judgment.
To elaborate, a correct caption—matching the human-annotated counterpart—should yield a substantial cosine similarity score between the generated and original images, as measured by our evaluation framework. Conversely, an incorrect caption—deviating from the human-annotated version—should result in a comparatively smaller cosine similarity score. 
This approach allows us to empirically validate the effectiveness of our proposed evaluation framework in aligning with human judgment.

\subsection{Experimental Settings}
To illustrate the application of the proposed framework for evaluating an image captioning model, we employ the InstructBLIP \cite{Dai2023InstructBLIPTG} model in our image captioning module. This model is equipped with the pre-trained language model Vicuna-7B \cite{Zheng2023JudgingLW} to generate image descriptions. Image captions are generated using the prompt ``<Image> A short image caption:'', guiding the model to produce sentences of fewer than $100$ tokens, excluding special symbols. 
For text-to-image generation, GPT-4 with the built-in diffusion model DALL-E-3 is employed. Notably, the diffusion model can be replaced by Stable Diffusion models \cite{Rombach_2022_CVPR}, utilizing a fixed, pre-trained encoder (ViT-g/14) \cite{Radford2021LearningTV}, and the entire diffusion model is pre-trained on the LAION-2B dataset \cite{Schuhmann2021LAION400MOD}.
Human evaluation serves as the validation method for the proposed framework. Each image in the dataset comes with five human-annotated image captions, and performance is quantified using the average cosine similarity score, as detailed in Section \ref{human:eva}. The experiments are conducted using two NVIDIA-A6000 GPUs.

\subsection{Datasets}
\label{dataset:mscoco_flickr}
\noindent\textbf{MSCOCO Dataset \cite{lin2014microsoft}.}
The MSCOCO dataset comprises two primary components: the images and their corresponding annotations. The images are organized into a directory hierarchy, with top-level directories for the train, validation, and test sets.
Annotations are provided in JSON format, with each file corresponding to a single image. Each annotation includes details such as the image file name, dimensions (width and height), a list of objects with their respective class labels (e.g., ``person,'' ``car''), bounding box coordinates ($x$, $y$, width, height), segmentation mask (in polygon or RLE format), keypoints and their positions (if available), and five captions describing the scene. Additional information provided by the MSCOCO dataset includes image super categories, license details, and coco-stuff annotations (pixel-wise annotations for stuff classes in addition to the $80$ object classes).
The MSCOCO dataset provides various types of annotations, including object detection with bounding box coordinates and full segmentation masks for $80$ different objects, stuff image segmentation with pixel maps displaying $91$ amorphous background areas, panoptic segmentation identifying items in images based on $80$ ``things'' and $91$ ``stuff'' categories, dense pose annotations featuring over $39,000$ photos and mapping between pixels and a template for over $56,000$ tagged persons, 3D model annotations and natural language descriptions for each image, and keypoint annotations for over $250,000$ persons annotated with key points such as the right eye, nose, and left hip.

\noindent\textbf{Flickr30k Dataset \cite{young2014image}.}
The authors in \cite{young2014image} advocate for utilizing the visual denotations of linguistic expressions, represented by the set of images they describe, to define new denotational similarity metrics. These metrics, as demonstrated in \cite{young2014image}, prove to be at least as advantageous as distributional similarities for tasks requiring semantic inference. The computation of these denotational similarities involves the construction of a denotation graph—a subsumption hierarchy over constituents and their denotations. This graph is established using a substantial corpus comprising $30,000$ images and $150,000$ descriptive captions.
The creation of this denotation graph involves the development of an image caption corpus by the authors in \cite{young2014image}, consisting of $158,915$ crowd-sourced captions elucidating $31,783$ images. This corpus serves as an extension of their previous work on the Flickr8k Dataset. The new images and captions specifically focus on individuals engaged in everyday activities and events.

\subsection{Effectiveness Analysis of the Proposed Evaluation Framework}

\noindent\textbf{Human Evaluation Using the Proposed Dataset.}
\label{human:eva}
The dataset introduced in this work, consisting of pairs of images and captions, has undergone human annotation. Each image is accompanied by five distinct human-generated captions. The details of our human evaluation process are outlined below. In Step 1, we directly utilize the human-annotated ground truth caption to generate an image through a text-to-image LLM, such as GPT-4 or Gemini. In Step 2, we extract the image features of both the ground truth caption's corresponding image and the image generated by the text-to-image LLM. In Step 3, we apply the cosine similarity formula from Section \ref{metric:cosine} to compute the cosine similarity scores between these two sets of image features.
Given that the caption is a human-annotated ground truth description, accurately portraying the corresponding image, we expect the similarity score from Step 3 to be high. Conversely, if a caption inaccurately describes a given image, the cosine similarity score from Step 3 should be low. Consistency between the experimental result and these expectations indicates the effectiveness of the proposed evaluation framework in aligning with human consensus.

\begin{figure}[t!]
\begin{center}
\centering
\includegraphics[width=1.0\linewidth]{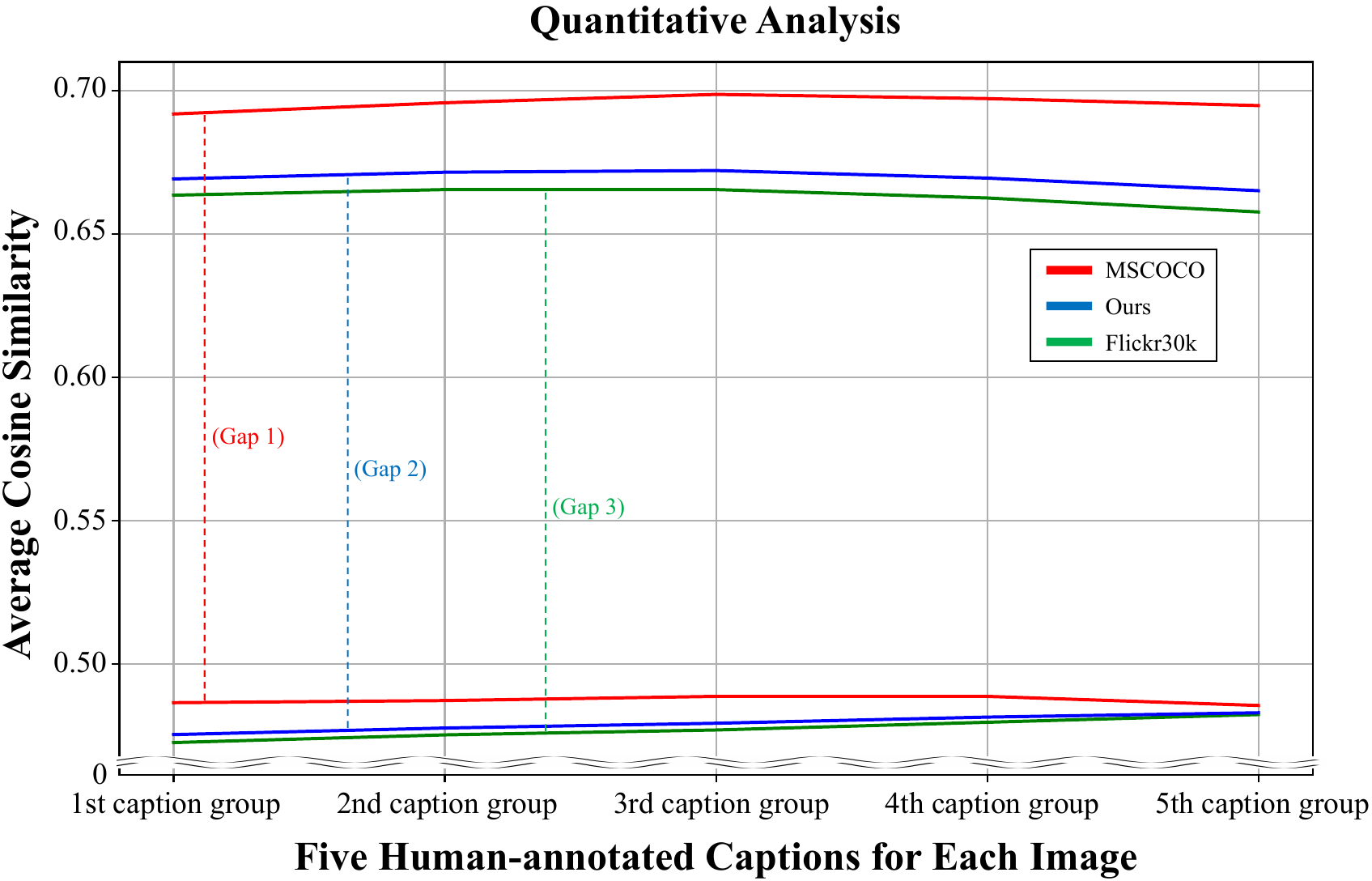}
\end{center}
\vspace{-0.4cm}
   \caption{Human evaluation results. The outcomes are derived from three datasets: MSCOCO (highlighted in red), Flickr30k (highlighted in green), and our dataset (highlighted in blue). The top three lines represent scenarios where the provided caption aligns with the correct human-annotated description, while the bottom three lines represent scenarios where the caption is incorrect. ``Gap 1'', ``Gap 2'', and ``Gap 3'' signify the disparities in average cosine similarity scores. We observe that these gaps are approximately $0.2$, indicating the influence of the provided captions on the cosine similarity scores. A larger gap indicates a substantial mismatch between the human-annotated image description and the provided or model-generated caption, whereas a smaller gap suggests a higher degree of alignment.
   }
\vspace{-0.6cm}
\label{fig:figure4}
\end{figure}

The evaluation results depicted in Figure \ref{fig:figure4} reveal notable insights. The blue lines in Figure \ref{fig:figure4} illustrate the impact of the provided captions on the cosine similarity scores. Specifically, when the provided caption matches the correct human-annotated description (upper blue line), the average cosine similarity score reaches approximately $0.67$. Conversely, when the caption is incorrect (lower blue line), the average cosine similarity score drops to around $0.47$. This discrepancy results in a similarity gap of approximately $0.2$. 
These findings underscore the effectiveness of the proposed evaluation framework, as it closely aligns with human judgment.
It is noteworthy that the robustness of this human evaluation method is attributed to the remarkable text-to-image generation capabilities of modern LLM models. Widely recognized models such as GPT-4 and Gemini have been extensively acclaimed in various studies and by the broader community \cite{brown2020language,team2023gemini,wu2023brief}.

\noindent\textbf{Assessment Using MSCOCO and Flickr30k Datasets.}
Figure \ref{fig:figure4} reveals consistent trends in the evaluation results across MSCOCO, Flickr30k, and our dataset. Similar patterns are observed in MSCOCO and Flickr30k, where there is a notable decrease in the average cosine similarity when the model-generated image caption differs from the human-annotated ground truth caption. These findings affirm the effectiveness and reliability of the proposed evaluation framework for assessing image captioning models.

\begin{figure*}[ht]
    \centering
    \begin{subfigure}[b]{0.45\textwidth}
        \centering
        \includegraphics[width=\textwidth]{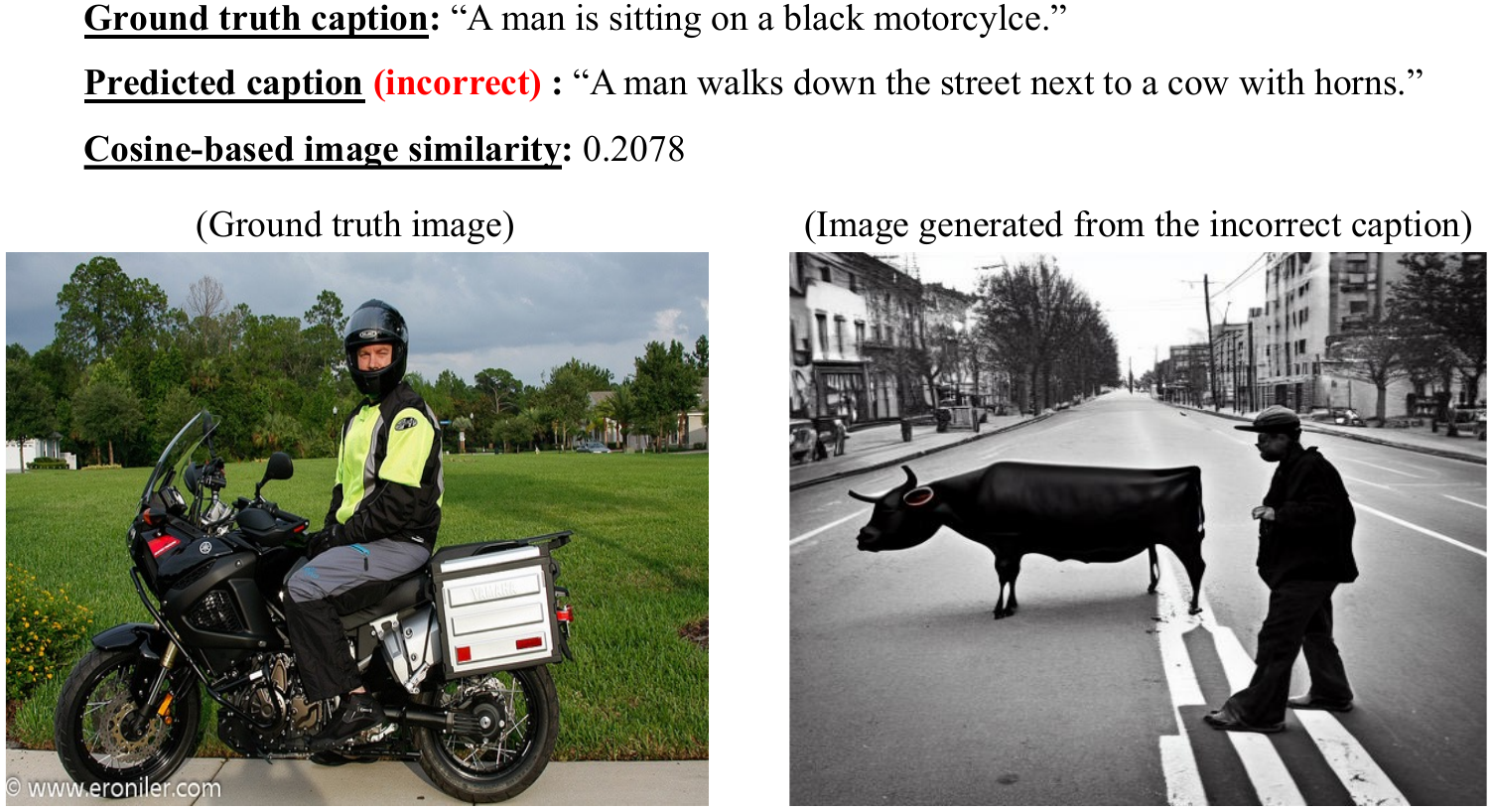}
        \caption{Example 1}
        \hspace{2cm} 
        \label{fig:sub3}
    \end{subfigure}
    \hfill
    \begin{subfigure}[b]{0.455\textwidth}
        \centering
        \includegraphics[width=\textwidth]{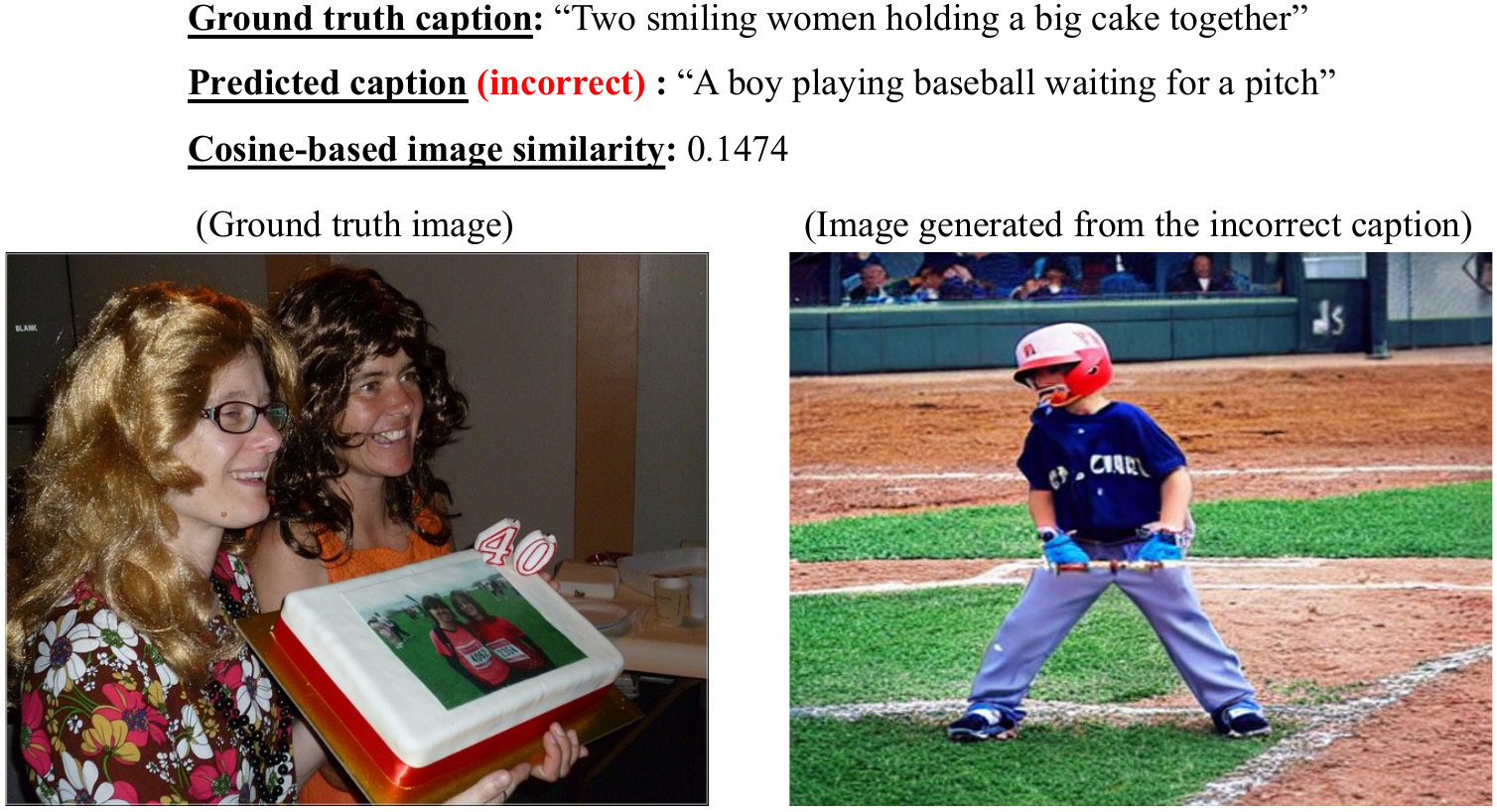}
        \caption{Example 2}
        \hspace{2cm} 
        \label{fig:sub4}
    \end{subfigure}
    \hfill
    \begin{subfigure}[b]{0.45\textwidth}
        \centering
        \includegraphics[width=\textwidth]{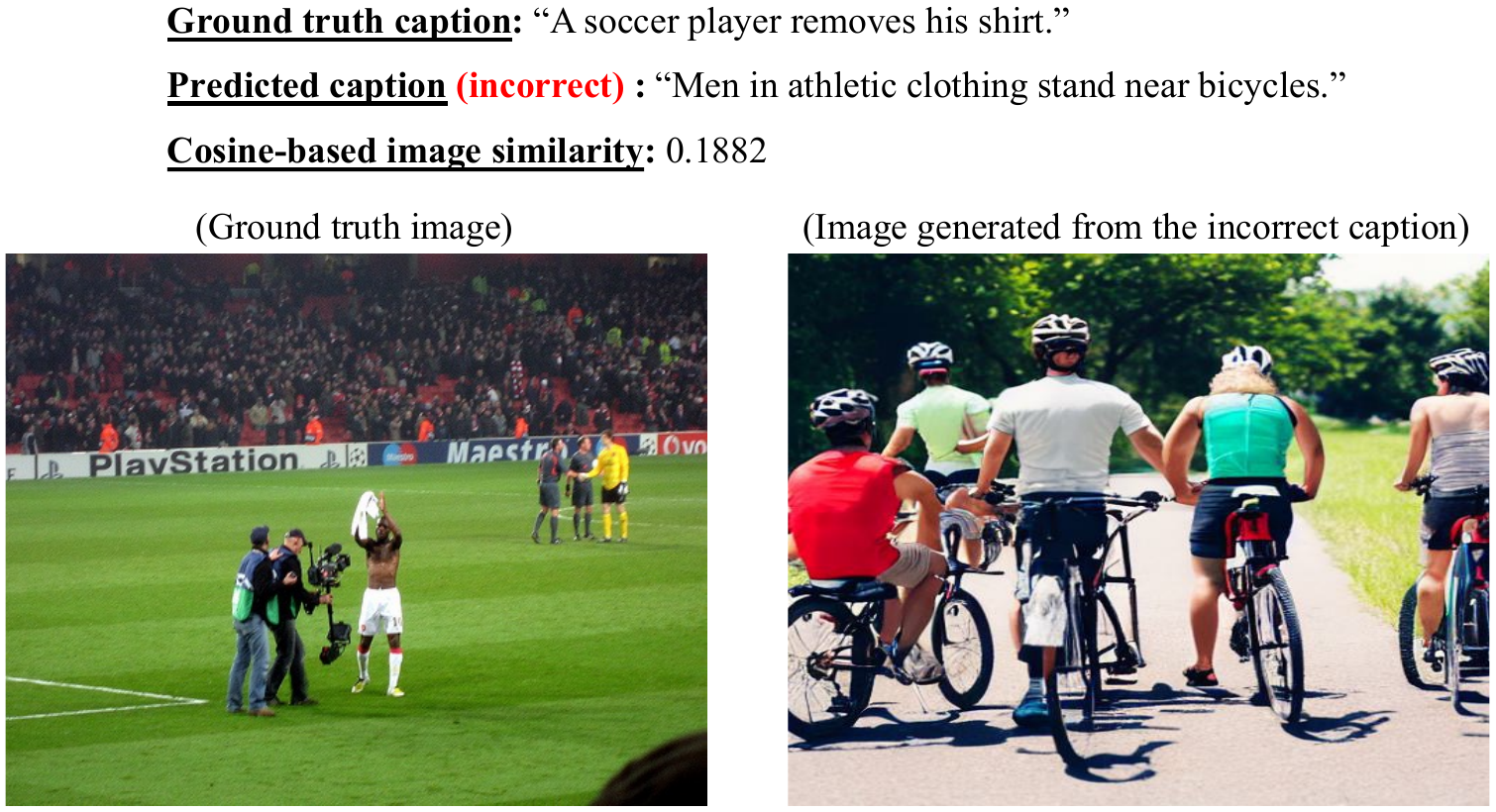}
        \caption{Example 3}
        \label{fig:sub5}
    \end{subfigure}
    \hfill
    \begin{subfigure}[b]{0.46\textwidth}
        \centering
        \includegraphics[width=\textwidth]{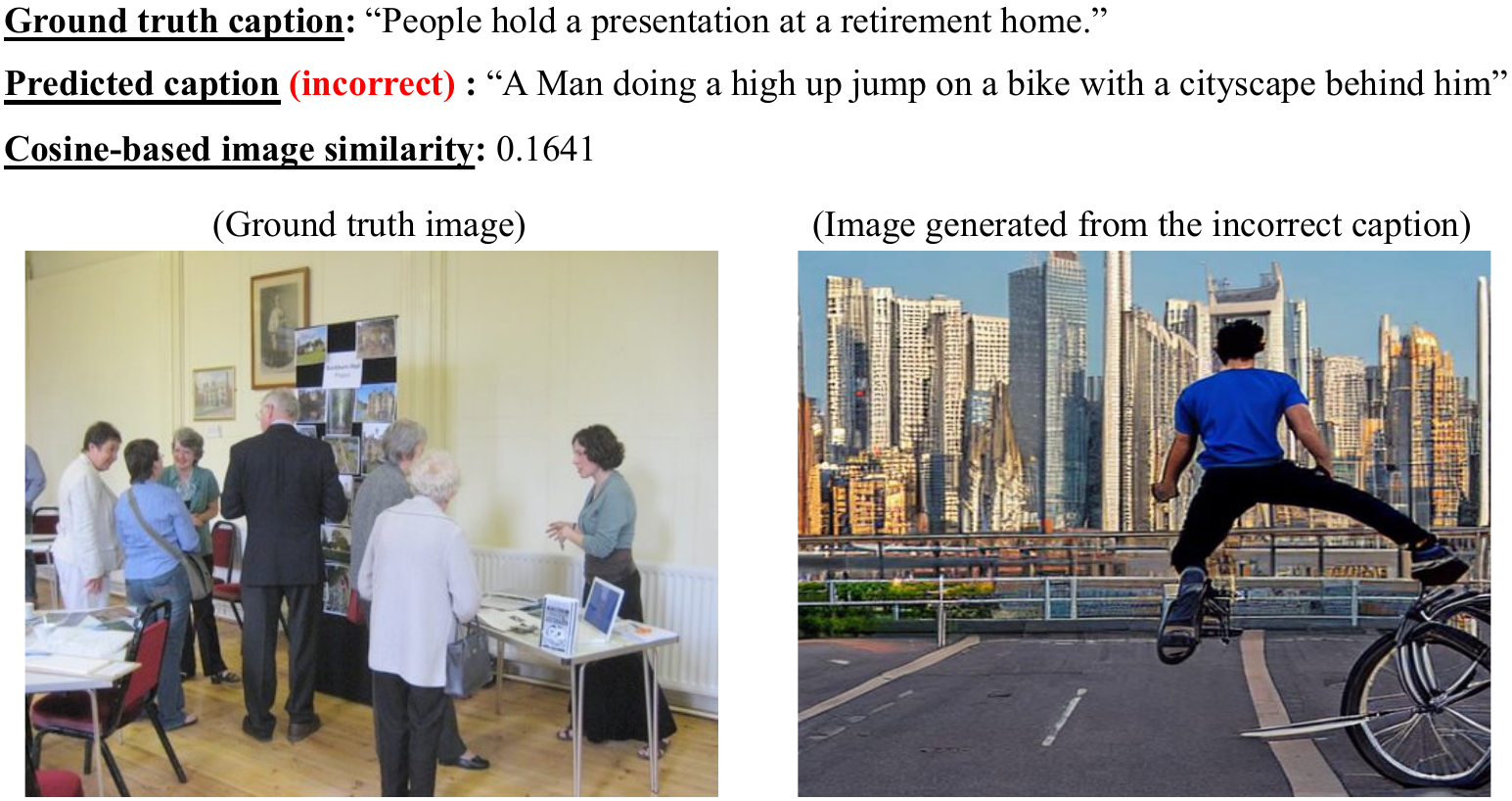}
        \caption{Example 4}
        \label{fig:sub6}
    \end{subfigure}
    \vspace{-0.3cm}
    \caption{Impact of incorrect image captions. Due to LLMs' proficiency in generating images accurately from provided text prompts, inconsistencies between model-generated image captions and human-annotated ground truth descriptions can lead to discrepancies in the generated images. Leveraging this observation, we propose an evaluation framework to assess the performance of image captioning models without using human-annotated ground truth captions.
    }
    \label{fig:figure7}
\end{figure*}

\begin{figure*}[ht]
    \centering
    \begin{subfigure}[b]{0.5\textwidth}
        \centering
        \includegraphics[width=\textwidth]{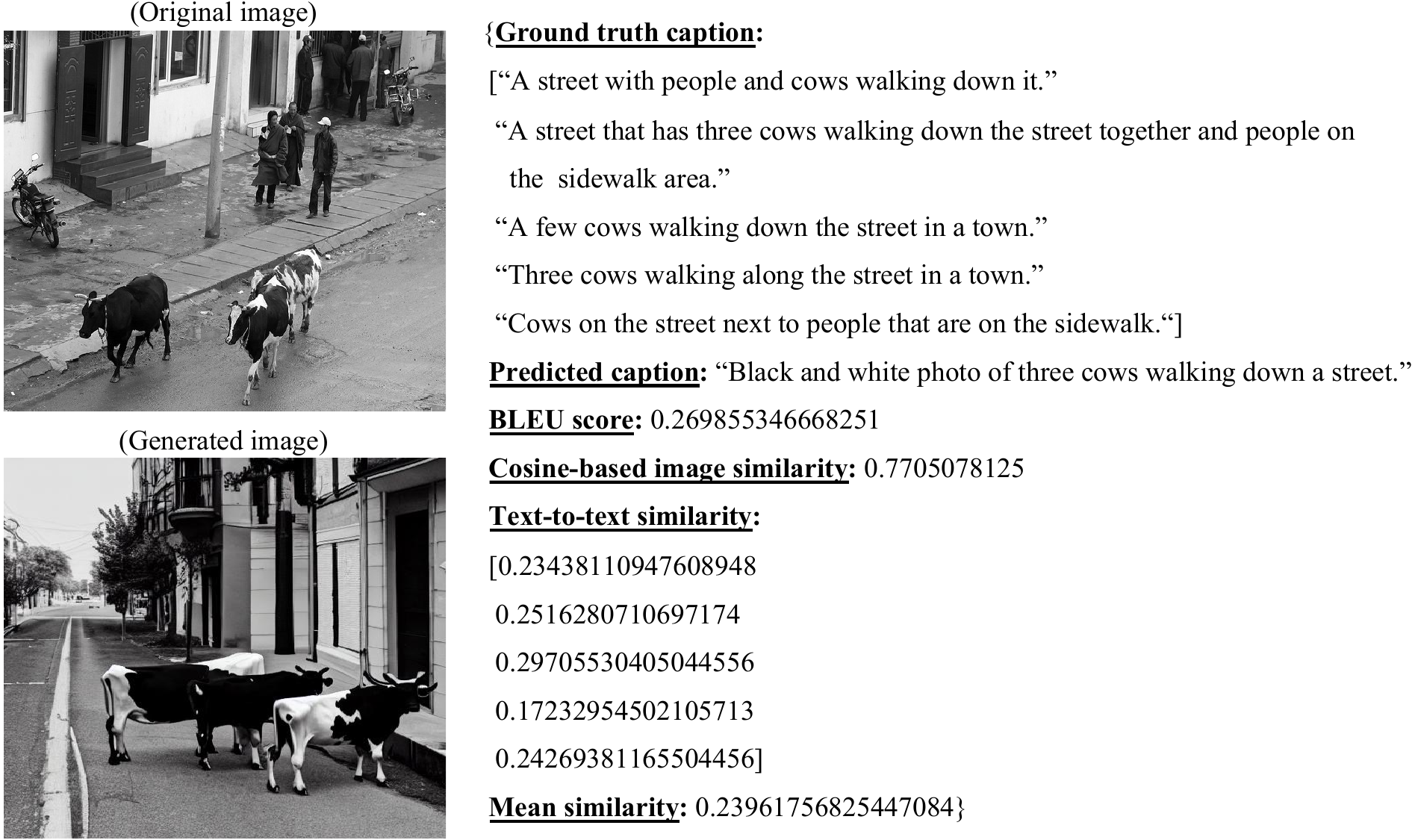}
        \caption{Example 1}
        \label{fig:sub1}
    \end{subfigure}
    \hfill
    \begin{subfigure}[b]{0.45\textwidth}
        \centering
        \includegraphics[width=\textwidth]{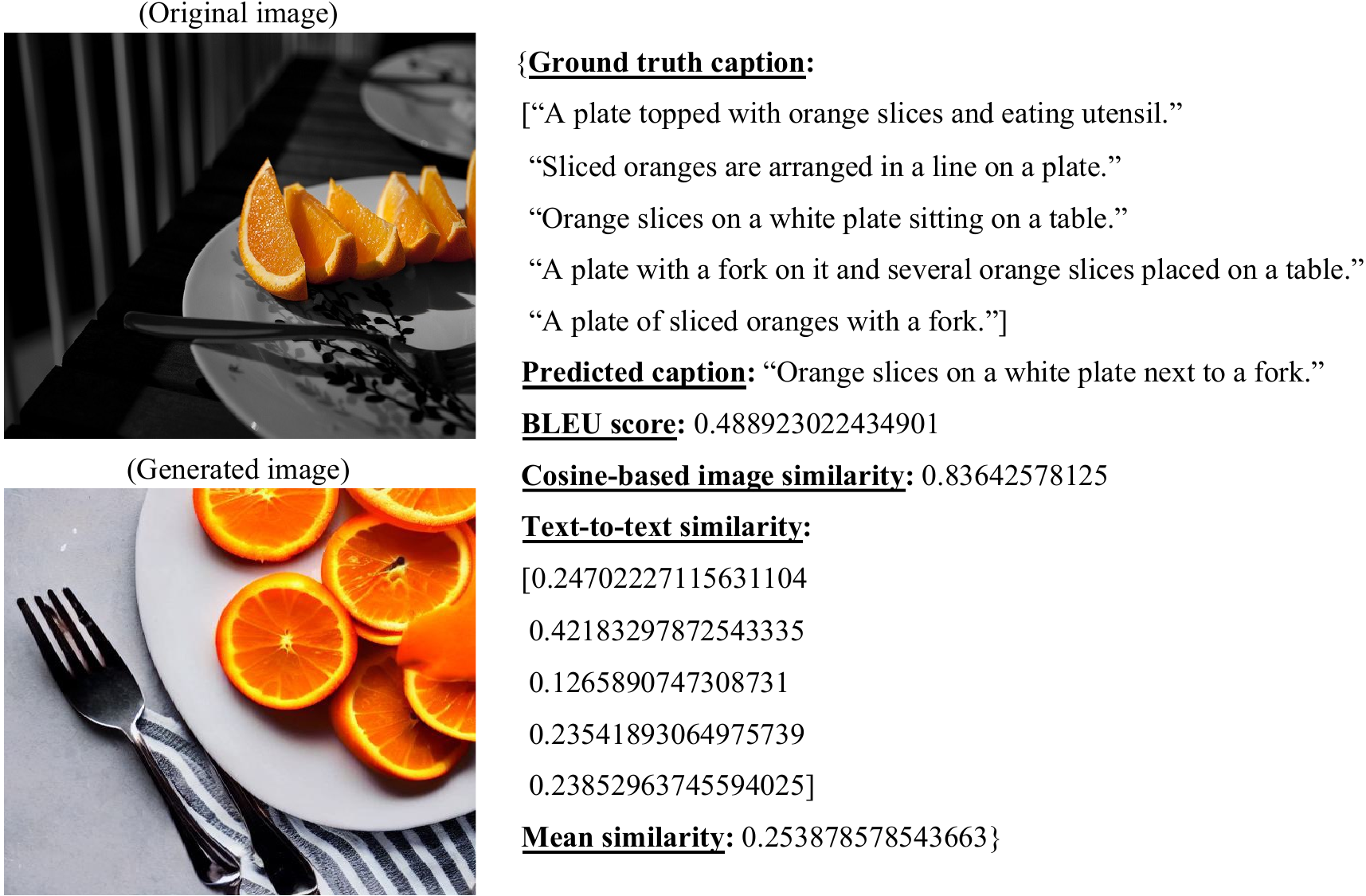}
        \caption{Example 2}
        \label{fig:sub2}
    \end{subfigure}
    \vspace{-0.3cm}
    \caption{Limitations of text-based evaluation metrics in image captioning. See Equation~(\ref{eq:bleu}) for the calculation of the BLEU score. ``Predicted caption'' refers to the caption generated by the InstructBLIP model. ``Text to text similarity'' indicates the cosine similarity between the human-annotated ground truth caption and the model-generated caption using text-based CLIP embeddings. ``Mean similarity'' represents the average of the five values of ``Text to text similarity''.}
    \label{fig:figure8}
    \vspace{-0.3cm}
\end{figure*}

\noindent\textbf{Qualitative Analysis.}
To gain deeper insights into the performance of the proposed evaluation framework, we present qualitative results in Figure \ref{fig:figure7} and Figure \ref{fig:figure8}. In Figure \ref{fig:figure7}, we observe that the human-annotated ground truth captions and the model-predicted captions exhibit poor alignment in these four examples. Given the accurate image generation capabilities of existing LLMs based on text-based prompts, the accuracy of model-generated image descriptions is crucial. However, in these instances, all predicted captions are incorrect, resulting in LLM-generated images that significantly differ from the ground truth images. Consequently, this discrepancy contributes to the low cosine-based similarity scores.

In Figure \ref{fig:figure8}, these two examples illustrate a strong alignment between the model-generated descriptions and the human-generated ground truth captions. Hence, this alignment results in LLM-generated images that closely resemble the ground truth images. As a result, when calculating cosine similarity scores based on the image features extracted from the LLM-generated and ground truth images, the scores are notably high.
We also calculate scores based on these metrics to highlight the advantage of our proposed method over the aforementioned text-based evaluation metrics. In Figure \ref{fig:figure8}, we observe that despite the model-generated image captions closely matching the ground truth captions, the scores based on text-based evaluation metrics are comparatively low. This observation underscores the superiority of our proposed evaluation framework over existing text-based evaluation metrics for image captioning models.
\begin{equation}
    \textup{BP}= \left\{\begin{matrix}
                        1 & \textup{if} & c>r \\ 
                        \textup{exp}(1-\frac{r}{c}) & \textup{if} & c\leq r
                        \end{matrix}\right.;
    \textup{BLEU}= \textup{BP}\cdot \textup{exp}\left ( \sum_{n=1}^{N} w_{n}\textup{log}p_{n} \right ),
    \label{eq:bleu}
\end{equation}
where $r$ represents the effective length of the ground truth text, $c$ signifies the length of the predicted text, and $\textup{BP}$ stands for brevity penalty. The geometric mean of the adjusted $n$-gram precisions $p_{n}$ is calculated using $n$-grams up to a length of $N$, with positive weights $w_{n}$ that sum to 1.

\section{Conclusion}
In this study, we have introduced a novel framework for evaluating automatically generated image descriptions, aiming to overcome the limitations of existing evaluation metrics like BLEU, ROUGE, METEOR, and CIDEr. Our framework leverages advancements in LLMs such as GPT-4 or Gemini to utilize image descriptions generated by an image captioning model for creating corresponding images. By quantifying the cosine similarity between the representation of the original input image in the image captioning model and the representation of the LLM-generated image, we can effectively assess the model's performance without relying on human-annotated reference captions.
Through extensive experiments on the established datasets like Flickr30k and MSCOCO, we have demonstrated the effectiveness of the proposed evaluation framework. Our experimental results suggest that the proposed framework's performance closely correlates with human judgment, offering a valuable method for evaluating the effectiveness of image captioning models. Additionally, human evaluations conducted on our introduced dataset validate the framework's efficacy in capturing various aspects such as grammaticality, coverage, correctness, and truthfulness in automatically generated image descriptions.
Moving forward, the proposed framework presents new opportunities for evaluating image captioning models, offering a more efficient and reliable alternative to traditional human evaluations and existing automated evaluation metrics. 
It is designed to complement, rather than replace, human judgment. In summary, our work contributes to the ongoing development of robust evaluation frameworks for image captioning models, bridging the gap between automated metrics and human judgment, and driving advancements in this field.
\bibliographystyle{ACM-Reference-Format}
\bibliography{sample-base}

%
\appendix

\end{document}